\documentclass[letterpaper, 10 pt, conference]{ieeeconf}
\IEEEoverridecommandlockouts  
\usepackage{cite}
\usepackage{amsmath,amssymb,amsfonts}
\usepackage{algorithmic}
\usepackage{graphicx}
\usepackage{textcomp}
\usepackage{xcolor}
\usepackage{mathrsfs}
\usepackage{amssymb}
\usepackage[flushleft]{threeparttable}
\usepackage{booktabs}
\usepackage[font=small]{caption}
\usepackage[font=small,position=b]{subcaption}

\usepackage{xparse} 
\usepackage{wrapfig}
\usepackage{hyperref}
\usepackage{cleveref}
\usepackage{dsfont}
\usepackage{comment}
\usepackage{color}
\usepackage{balance}

\NewDocumentCommand\bbm{}{ \begin{bmatrix} }
	\NewDocumentCommand\ebm{}{ \end{bmatrix} }

\NewDocumentCommand\Matrix{m}{ \boldsymbol{\mathbf{#1}} }

\NewDocumentCommand\LieGroupSO{m}{ \mathrm{SO}(#1) }

\NewDocumentCommand\LieGroupSE{m}{ \mathrm{SE}(#1) }
\NewDocumentCommand\LieAlgebraSE{m}{ \mathfrak{se}(#1) }

\NewDocumentCommand\AbsoluteValue{m}{ \left\vert#1\right\vert }
\newcommand{\norm}[1]{\left\lVert#1\right\rVert}

\NewDocumentCommand\Rotation{}{ \Matrix{R} }

\NewDocumentCommand\Transform{}{ \Matrix{T} }

\NewDocumentCommand\MatExp{m}{\mathrm{Exp}\left({#1}\right)}
\NewDocumentCommand\Matlog{m}{\mathrm{log}\left({#1}\right)^{\vee}}

\NewDocumentCommand\DPCNet{}{\texttt{DPC-Net}}

\begin{document}
\title{Self-Supervised Deep Pose Corrections for Robust Visual Odometry}
\author{Brandon Wagstaff, Valentin Peretroukhin, and Jonathan Kelly
\thanks{All authors are with the Space \& Terrestrial Autonomous Robotic Systems (STARS) Laboratory at the University of Toronto Institute for Aerospace Studies (UTIAS), Toronto, Ontario, Canada, M3H~5T6. Email: \texttt{<first name>.<last name>@robotics.utias.utoronto.ca}}
}
\maketitle
\thispagestyle{empty}
\pagestyle{empty}
\begin{abstract}
We present a self-supervised deep pose correction (DPC) network that applies pose corrections to a visual odometry estimator to improve its accuracy. Instead of regressing inter-frame pose changes directly, we build on prior work that uses data-driven learning to regress pose corrections that account for systematic errors due to violations of modelling assumptions. Our self-supervised formulation removes any requirement for six-degrees-of-freedom ground truth and, in contrast to expectations, often improves overall navigation accuracy compared to a supervised approach. Through extensive experiments, we show that our self-supervised DPC network can significantly enhance the performance of classical monocular and stereo odometry estimators and substantially out-performs state-of-the-art learning-only approaches.
\end{abstract}

\section{Introduction}

Accurate self-localization is a prerequisite for reliable mobile autonomy and is especially important in situations where global navigation satellite system signals are unavailable or unreliable. Vision-based self-localization, in particular, has become ubiquitous since high-quality cameras are now	 relatively inexpensive and compact. Despite having a rich history in computer vision and robotics \cite{Aqel:2016}, visual localization still remains an open research topic, particularly in dynamic environments where many common modelling assumptions are violated \cite{Cadena:2016}.

At the heart of visual self-localization is visual odometry (VO): the process of estimating a camera's own motion (or \textit{egomotion}) from sequential image captures. To remain computationally tractable, `classical' VO algorithms have typically assumed that the scene consists of static objects, has constant illumination, and lacks major occlusions. However, for long-term autonomy, it is critical for VO-based algorithms to maintain accuracy in spite of such adverse effects. Recently, in hopes of achieving robust pose estimates, end-to-end learning-based approaches have been proposed that completely replace classical techniques with learned models. By learning directly from data (using supervised \cite{Wang:2017} or self-supervised \cite{Li:2018} methods), these network-based techniques have the potential to relax many of the assumptions that classical VO pipelines make, and as a result, to be robust to moving objects, poor illumination, and significant occlusions. However, to date, end-to-end-approaches have not surpassed the accuracy of classical VO algorithms.

Alternatively, other methods have augmented (rather than replaced) classical estimators with learned components. For example, learned measurement models have been used within Kalman filters \cite{Haarnoja:2016} as a way to extract illumination direction from monocular images in an effort to reduce orientation drift \cite{2017_Peretroukhin_Reducing}, or as a way to more accurately initialize depth within a monocular VO pipeline \cite{Yang:2018DeepVS}. By combining learning with classical pipelines, these methods aim to retain the interpretability and transferability of model-based techniques while leveraging the capacity and flexibility of data-driven model-free learning to improve accuracy and robustness. 

\begin{figure}[t]
	\centering
	\includegraphics[width=0.97\columnwidth]{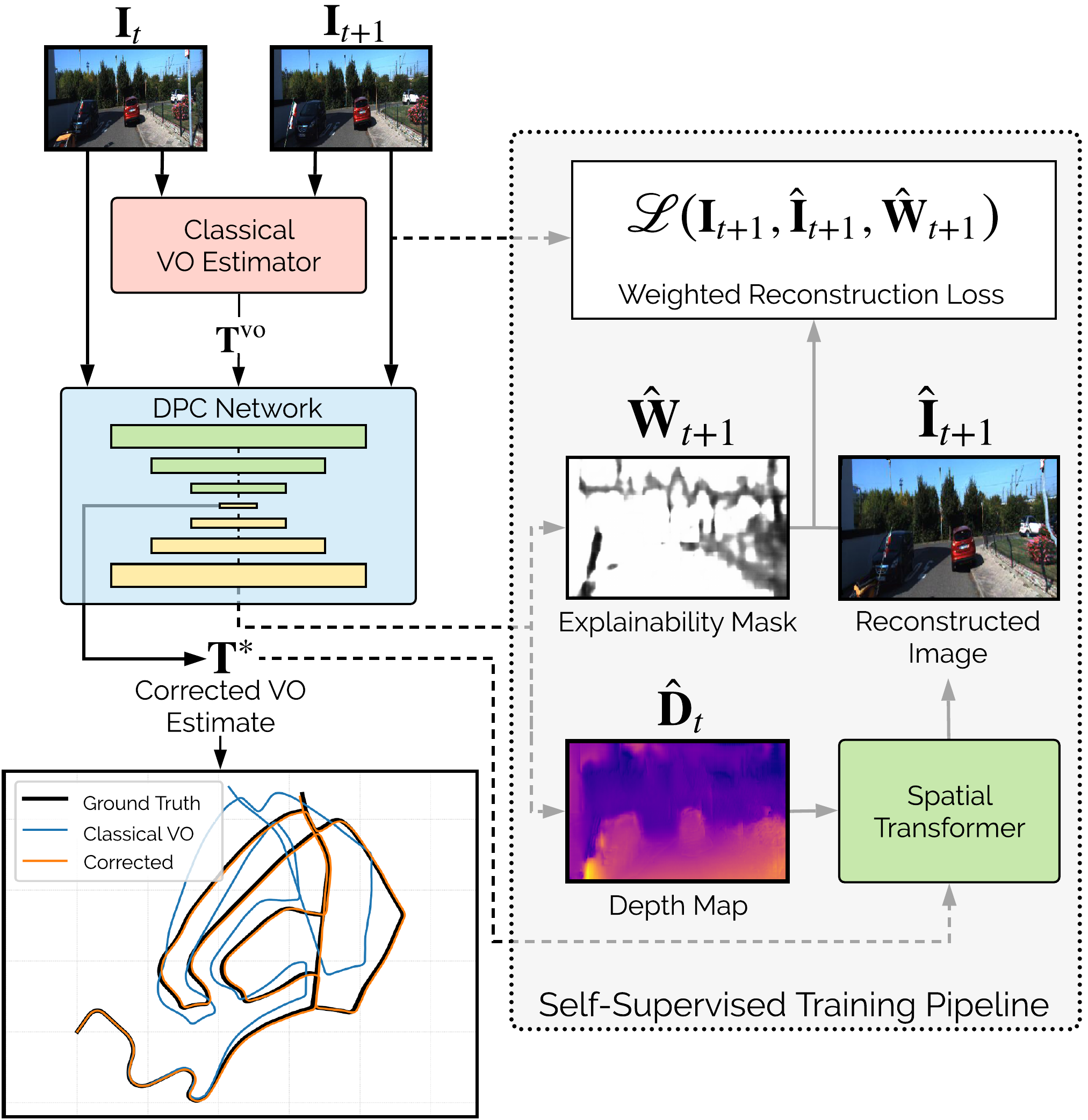}
	\vspace{1mm}
	\caption{Our self-supervised deep pose correction (DPC) network regresses a pose correction to a classical VO estimator.}
	\label{fig:overview}
	\vspace{-0.65cm}
\end{figure}

Our approach, illustrated in \Cref{fig:overview}, follows in the spirit of these latter techniques. We fuse a classical VO pipeline with a data-driven model through the paradigm of learned pose \textit{corrections}. Rather than training a network to regress the full inter-frame pose change from data alone, we instead rely on a classical VO estimate to produce a `large' prior and use a deep neural network (DNN) to learn a smaller `correction' that models how this classical estimator degrades in adverse situations (e.g., when there are many dynamic objects within the scene). We extend \DPCNet{}, the system proposed by \cite{Peretroukhin:2018}, which introduced the paradigm of  deep pose corrections (DPC) and used a DNN to predict corrections through a supervised training method driven by ground truth pose information. We improve upon this by replacing the supervised pose loss with a self-supervised photometric reconstruction loss that obviates the need for ground truth pose labels, which are often expensive or impractical to collect. Our loss formulation facilitates continual retraining of the network when traversing new environments, since no supervision is required for training. In short, our novel contributions are:

\begin{enumerate}
	\item a deep pose correction network that can be trained in a self-supervised manner, which obviates the need for ground truth pose labels,
	\item substantial experimental validation of our method on the KITTI odometry dataset, showing that our method achieves state-of-the-art accuracy, and
	\item the release of an open source implementation of our method in \texttt{PyTorch} \cite{paszke:2017}.\footnote{See \url{https://github.com/utiasSTARS/ss-dpc-net}}
\end{enumerate}

\section{Background}

Visual odometry is a well studied navigation technique that is used to estimate the robot's six-degrees-of-freedom (DoF) pose change by solving for the camera's egomotion between image frames. Typically, the inter-frame pose change is determined by optimizing for the pose that minimizes the error when aligning two sets of 3D points (for indirect methods) or pixels intensities (for direct methods) in a sequence of monocular or stereo camera images. Outlier rejection methods such as RANSAC \cite{Fischler:1981} or robust losses are used to remove (or downweight) features (or pixels) that adversely affect the optimization process. By compounding relative pose changes, a global pose estimate is determined. However, VO is subject to superlinear error growth: in general, any small misestimate of the relative orientation change leads to larger and larger position errors. Typically, sources of error for VO include false correspondences (during feature matching), poor feature detection (due to camera motion blur or poor lighting), or the presence of dynamic objects within the scene. We refer the reader to \cite{Scaramuzza:2011} for a detailed and comprehensive review of visual odometry.

New data-driven paradigms for VO replace portions (or all) of the classical localization pipeline with a learned model. Some approaches \cite{Wang:2017} use supervised learning to train a convolutional neural network (CNN) to regress inter-frame pose changes in an end-to-end manner. Other techniques \cite{zhou:2017,Prasad:2018,Vijayanarasimhan:2017} rely on a self-supervised photometric reconstruction loss formulation for end-to-end VO. These methods train a network to regress the relative pose change between a current (source) view and a nearby (target) view by minimizing a photometric reconstruction loss. Such loss functions penalize the differences between the pixel intensities of the target image and the reconstructed image: under the assumptions that the scene is static, has constant illumination, and that there are no occlusions, the target image is reconstructed from the source image through an inverse compositional warping procedure that uses the predicted inter-frame pose change, a depth map (which is also learned) and the known camera intrinsic parameters. To train this type of network, a differentiable image warping tool called a spatial transformer \cite{Jaderberg:2015} is applied to efficiently synthesize the reconstructed image; this allows gradients to be backpropagated from the reconstruction loss. Our work extends the self-supervised pipeline to learn pose corrections, instead of full poses, based on a similar photometric loss.

Several others systems build on the baseline photometric reconstruction loss by imposing additional constraints. The authors of \cite{Iyer:2018} use ``composite transform constraints'' to ensure that the predicted pose change across multiple frames is similar to the pose change produced by compounding the predicted pose changes between each individual image pair. In \cite{Prasad:2018}, epipolar geometry constraints are enforced to ensure that pixels from the source image are reprojected near to the epipolar line in the target image. The system in \cite{Li:2018} uses a left-right stereo consistency loss that allows the depth network to output scaled depth maps from monocular camera images at test time. The approach described in \cite{Zhu:2018} operates by learning to predict optical flow and disparity and then uses a classical RANSAC outlier rejection scheme to select a set of inlying pixels that can be used for pose estimation. The authors in \cite{zhou:2017} train a network to additionally regress an ``explainability mask,'' which ignores unreliable pixels that hinder image reconstruction, either because they break the photometric consistency assumption or because they correspond to objects that are moving. Our network relies on the explainability mask defined in \cite{zhou:2017}.

\section{Approach}
Our approach (\Cref{fig:overview}) merges the self-supervised training procedure of end-to-end VO networks with the DPC framework of \cite{2017_Peretroukhin_Reducing}. We replace the supervised loss of \DPCNet{} with a photometric reconstruction loss that does not require any external ground truth pose information, yet can still produce accurate pose corrections. Similar to other self-supervised methods, our network outputs a predicted inter-frame pose change and a depth map---however, our predicted inter-frame pose change is `initialized' with an egomotion \textit{prior} from a classical VO estimator. We compound this prior with a \textit{correction} that is produced by our pose network. Using the depth map and the corrected inter-frame pose, we warp a source image into a target image and evaluate a photometric reconstruction loss. Unlike \cite{2017_Peretroukhin_Reducing}, which uses a supervised pose loss and thus requires $\LieGroupSE{3}$ labels for training, our self-supervised photometric loss obviates the need for this type of 6-DoF ground truth, which can often be arduous to obtain.

Concretely, instead of directly estimating the inter-frame pose change, $\mathbf{T}_{t+1,t}$, our pose network aims to regress an $\LieGroupSE{3}$ \textit{correction}, $\mathbf{T}_{t+1,t}^\text{corr}$, that corrects a classical	 VO estimate, $\mathbf{T}_{t+1,t}^\text{vo}$,
\begin{align}
\mathbf{T}_{t+1,t}^* = \mathbf{T}_{t+1,t}^\text{corr}\mathbf{T}_{t+1,t}^\text{vo}. 
\end{align}
To parameterize this correction, we use an unconstrained vector from the $\LieAlgebraSE{3}$ Lie algebra, $\boldsymbol{\xi}^\text{corr}_{\mathrm{t+1,t}} \in \mathds{R}^{6\times1}$, and then apply the (capitalized) exponential map to produce an on-manifold $\LieGroupSE{3}$ correction:\footnote{Our notation is based on and consistent with \cite{Barfoot:2017,Sola:2018}, where a detailed review of matrix Lie groups is found.}
\begin{align}
\mathbf{T}_{t+1,t}^\text{corr} = \MatExp{\boldsymbol{\xi}^\text{corr}_{t+1,t}}.
\end{align}

Our network can be paired with any classical VO estimator and then trained to produce pose corrections specific to that estimator; inter-frame pose changes from the VO estimator are acquired for each pair of frames in the training dataset, and are then used to train the network to regress pose corrections that will minimize the photometric reconstruction loss. Herein, we train our system with a monocular and a stereo VO estimator (\texttt{libviso2-m} and \texttt{libviso2-s} \cite{Geiger:2011}, respectively) and show that our approach improves the localization accuracy of both estimators. We apply \textit{monocular} corrections (our DPC network only takes as input images from a single camera) to \texttt{libviso2-m} and \texttt{libviso2-s}, as our loss formulation does not enforce any stereo image constraints; we leave this possibility as future work.

\subsection{Image Warping Function}

We apply an inverse compositional warping function that uses the source image's estimated depth map, $\hat{\mathbf{D}}_{t}$, the camera intrinsics, and the (estimated) corrected pose change between frames. Assuming a pinhole camera model, image coordinates $\mathbf{u}_t = \begin{bmatrix}u_t & v_t \end{bmatrix}^T$ correspond to a 3D point $\mathbf{p}_{t}(\mathbf{u}_t) = \begin{bmatrix} x_t&y_t&z_t\end{bmatrix}^T$ in the scene:
\begin{align}
\mathbf{p}_{t}(\mathbf{u}_t) &=\hat{\mathbf{D}}_{t}(\mathbf{u}_t)\begin{bmatrix} \frac{u_t - c_u}{f_u} & \frac{v_t - c_v}{f_v} & 1 \end{bmatrix}^T,
\end{align}
where ($c_u$, $c_v$) is the camera's principal point, and ($f_u$, $f_v$) are the camera focal lengths in the horizontal and vertical directions, respectively. The estimated pose change between images is used to transform $\mathbf{p}_{t}(\mathbf{u}_t)$ to its 3D position at the next time-step,
\begin{align}
\hat{\mathbf{p}}_{{t+1}}(\mathbf{u}_t) = \mathbf{T}_{t+1,t}^*\,\mathbf{p}_{{t}}(\mathbf{u}_t).
\end{align}
The 3D coordinates $\hat{\mathbf{p}}_{{t+1}}(\mathbf{u}_t)$ are reprojected onto the image plane according to
\vspace{-0.05cm}
\begin{align}
\begin{bmatrix}\hat{\mathbf{u}}^{T}_{t+1} &\!\! 1 \end{bmatrix}^{T} &= \begin{bmatrix} f_u & 0 & c_u \\ 0 & f_v & c_u \\ 0 & 0 & 1 \end{bmatrix} \frac{1}{\hat{z}_{t+1}}\hat{\mathbf{p}}_{{t+1}}(\mathbf{u}_t),
\end{align}
and the reconstructed target image is repopulated at the predicted pixel location with the original pixel intensity from the source image, $\hat{\mathbf{I}}_{t+1}(\hat{\mathbf{u}}_{t+1}) = \mathbf{I}_{t}(\mathbf{u}_t)$.
Instead of reconstructing every pixel individually, we use a spatial transformer ($ST$) \cite{Jaderberg:2015} to perform differentiable image warping, which efficiently reconstructs the entire target image from the source image:
\begin{align}
\hat{\mathbf{I}}_{{t+1}} = ST(\mathbf{I}_{t}, \hat{\mathbf{D}}_{t}, \mathbf{T}_{t+1,t}^*, f_u, f_v, c_u, c_v).
\end{align}

\subsection{Loss Function}

\begin{figure*}[t]
	\centering
	\begin{subfigure}[]{1.4\columnwidth}
		\includegraphics[width=\columnwidth]{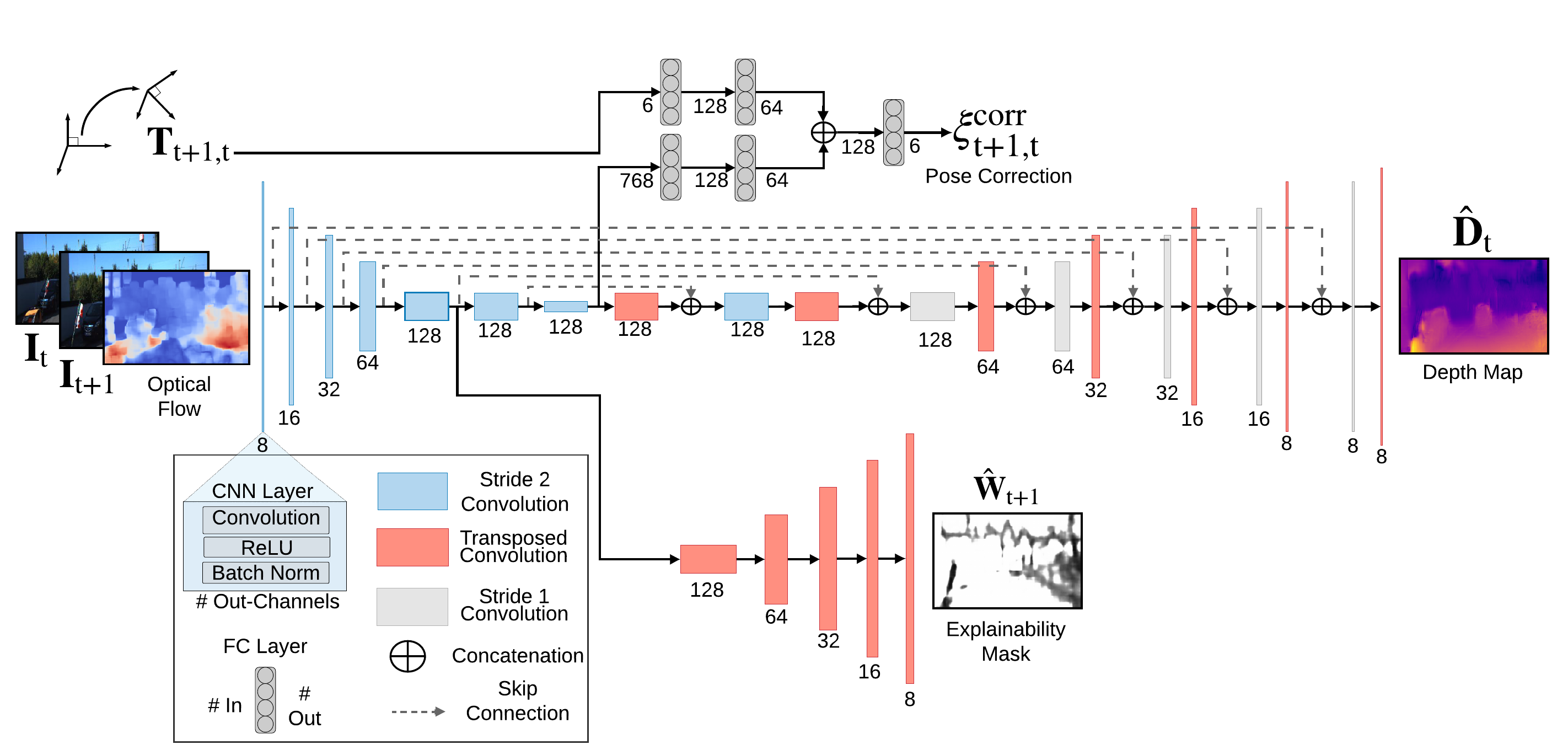}
		\caption{Our network jointly estimates a pose correction, a depth map, and an explainability mask.}
		\label{fig:network}
	\end{subfigure}
	\hfill
	\begin{subfigure}[]{0.56\columnwidth}
		\includegraphics[width=\columnwidth]{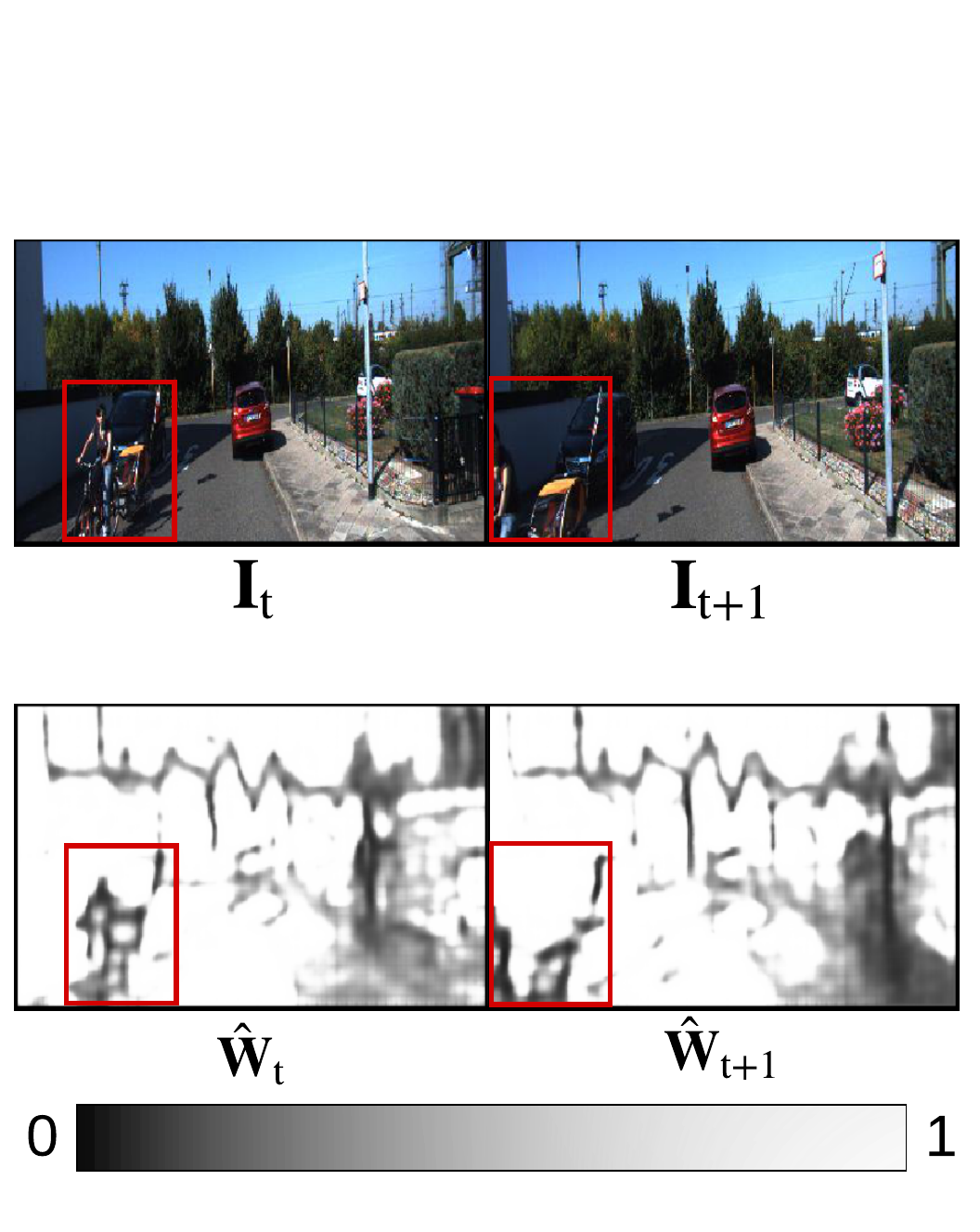}
		\caption{The explainability mask downweights unstable pixels.}
		\label{fig:expmask}
	\end{subfigure}
	\caption{A network diagram of our self-supervised DPC network and an example of where its explainability mask is useful.}
	\vspace{-0.2cm}
\end{figure*}

We use a weighted photometric reconstruction loss, which compares the reconstructed image with the target image. For a dataset with N training examples (each consisting of a source and target image of dimension $H\times W$ with $C$ colour channels), we define our loss function as:\footnote{Note that while many similar implementations use a multi-scale loss, we found that a multi-scale loss was detrimental to the quality of our pose corrections; we suspect this is the case because the regressed pose corrections are very small quantities, which require a high image resolution to learn.}
\begin{align}
\label{eq:loss}
\mathscr{L} &= \frac{1}{NCHW}\sum_{n=1}^N \sum_{u,v} \left( \mathscr{L}_\text{phot} + \lambda_\text{exp}\mathscr{L}_\text{exp}. + \lambda_\text{rot}\mathscr{L}_\text{rot} \right).
\end{align}
We now describe each of the three terms of this loss in detail. The first term, the pixel-wise weighted photometric reconstruction loss, compares a pixel $(u,v)$ from the target image with the corresponding pixel from the reconstructed image:
\begin{align}
\mathscr{L}_\text{phot} = \hat{\mathbf{W}}_{{t+1}}(u,v)\left| \hat{\mathbf{I}}_{{t+1}}(u,v) - \mathbf{I}_{{t+1}}(u,v) \right|.
\end{align}
Each pixel loss is weighted by the explainability mask, $\hat{\mathbf{W}}_{{t+1}}(u,v) \in (0,1)$, which accounts for situations in which the photometric consistency approximation is violated (e.g., due to lighting changes, dynamic objects, or occlusions). A high quality image reconstruction generally implies that the network's depth and pose correction estimates are also of high quality. To prevent the trivial solution of setting all explainability weights to zero, we use a regularization term, $\mathscr{L}_\text{exp}$, that is a cross-entropy loss with a constant label 1 for each pixel, 
\vspace{-0.1cm}
\begin{align}
\mathscr{L}_\text{exp} = -\log \hat{\mathbf{W}}_{{t+1}}(u,v).
\end{align}
\Cref{fig:expmask} illustrates an example scene where this mask is especially useful for mitigating the effects of moving objects. 

Lastly, $\mathscr{L}_\text{rot}$ is a loss term for training samples that incorporate large rotations; since these samples are less common than samples with smaller rotations, but are significantly more important for egomotion estimation, we use this loss to increase their relative weight compared to other samples. The loss term $\mathscr{L}_\text{rot}$ is the same as the photometric reconstruction term, but is set to zero for all samples except those with large rotations (according to the classical estimator's orientation estimate, $\Rotation_\text{vo}$):

\vspace{-0.2cm}
\begin{align} \mathscr{L}_\text{rot} &= \begin{cases} 
0 & \norm{\Matlog{\Rotation_\text{vo}}} < \gamma \\
\mathscr{L}_\text{phot} & \norm{\Matlog{\Rotation_\text{vo}}} \geq \gamma \\
\end{cases}
\end{align} \vspace{-0.2cm}

\subsection{Model Architecture}

Our network jointly estimates the pose correction, depth prediction, and explainability mask. \Cref{fig:network} provides a graphical illustration of the network structure, which is a modified version of the network in \cite{zhou:2017} that used a U-Net encoder-decoder \cite{Ronneberger:2015}. The network inputs are two images (the source and the target image) concatenated with the optical flow vectors estimated between the two images. The optical flow vectors, generated using the Gunnar-Farneb\"{a}ck algorithm \cite{Farneback:2003}, are incorporated because it has been shown in \cite{Zhou:2019-cv} that using intermediate representations adds explicit knowledge to the network that can improve performance on vision-based tasks. Additionally, we provide the network with the classical VO estimate (parameterized as a $6\times1$ Lie algebra vector through the logarithmic map), which is concatenated with the fully-connected layers near the centre of the network. Notably, we depart from prior work by unifying the depth and pose networks into a single network, which significantly improves the ability to generalize beyond the training data; incorporating the optical flow and the \texttt{libviso2} pose estimates as inputs additionally improved the results. 

The encoder network is composed of five blocks (in blue); each consists of a (stride 2) 2D convolution layer, a ReLU activation, and a batch normalization layer. We avoid the use of pooling layers, as they lead to spatial invariance, which would be detrimental for visual localization tasks. At the bottleneck, we branch the network into depth, explainability mask, and pose correction subnetworks. For the depth and explainability mask, we upsample from the bottleneck using decoder blocks which consist of a 2D transposed convolution layer \cite{dumoulin:2016} followed by a ReLU activation. Our depth prediction layer is a 2D (stride 1) convolution that reduces the channel layer to one, and a ReLU activation which ensures the output is positive.\footnote{In practice, we output the inverse depth estimate (to avoid regressing to infinite depths), and invert the network output to produce a depth estimate. Our network actually outputs a depth prediction at each of the final convolutional layers; each prediction is concatenated with the next layer (e.g., the prediction from the fourth-last layer is concatenated with its other output channels prior to being passed to the third-last layer). For interpretability, we omitted this from the network diagram.} The final explainability mask layer is a sigmoid activation, which compresses the pixel values to lie within $(0,1)$.

\vspace{-0.1cm}
\section{Experiments}

To train, validate, and test our system, we used the KITTI odometry dataset \cite{Geiger:2012,Geiger:2013}. Our full dataset consisted of colour images from the odometry sequences \texttt{00}, \texttt{02}, \texttt{05}-\texttt{10}, and an additional 24 training sequences (approximately 10,000 images) from the ``city'', ``residential'', and ``road'' categories of the raw KITTI dataset. The images were preprocessed to be more amenable for training: they were resized to $240 \times 376$ pixels and whitened using the ImageNet \cite{Deng2009-bx} statistics. Furthermore, similar to keyframe-based approaches, we used the classical VO estimates to filter out frames from the sequence that had little motion by removing those frames whose inter-frame translation or rotation was less than $1.5$ m or $0.4^{\circ}$, respectively. For all of the sequences, we estimated the camera poses using the \texttt{libviso2} package \cite{Geiger:2011}, which is a popular classical VO estimator (and, for consistency, is the same estimator used in \DPCNet{} \cite{Peretroukhin:2018}). We generated monocular (\texttt{libviso2-m}) and stereo (\texttt{libviso2-s}) egomotion estimates; these estimates are included in our open source repository. 

DPC models were trained for both the monocular and stereo \texttt{libviso2} estimators. We trained the DPC networks for up to 30 epochs with the Adam optimizer \cite{Kingma:2014} (with minibatch sizes of 32) using an initial learning rate of $1\times10^{-3}$ and $5\times10^{-5}$ for the stereo and monocular DPC models, respectively. We reduced the learning rate by a factor of 0.5 every ten and four epochs for the monocular and stereo DPC models, respectively. We used dropout ($p=0.5$) for all of the fully connected layers with a weight decay coefficient of $4\times10^{-6}$. All other hyperparameters were held constant during training. In our loss function (\Cref{eq:loss}), we selected the hyperparameter $\lambda_\text{exp}=0.23$ to cause the explainability mask to typically regress values close to one; progressively larger weights caused the explainability mask outputs to uniformly decrease, rather than decreasing only for unreliable pixels. We selected $\lambda_\text{rot}=4$ and $\gamma=0.005$ in order to emphasize larger rotations in the cost function. Using leave-one-out cross-validation, we trained a unique model for each test sequence, while using a single sequence for validation and all other sequences for training.

The primary challenge during training was selecting the training epoch whose model parameters produced high quality pose corrections; since no ground truth pose information was used in the training procedure, the network learned to minimize the photometric reconstruction error, not the localization error. Consequently, although the training procedure generally resulted in high-accuracy pose corrections, there were epochs that resulted in a low validation loss, but not in high-quality pose corrections. To address this, we developed two criteria to identify the training epoch with the most accurate pose corrections---both are evaluated in \Cref{sec:results}. 

\textit{Gradient Criterion}: First, we recompute the photometric loss \textit{only} for pixels with large gradient values (since the photometric errors for these pixels are highly sensitive to an erroneous pose correction). To compute the gradient loss, we generate a gradient mask by filtering out all pixels whose gradient value is below a constant, $\gamma_\text{grad}$:
\begin{align}
\frac{\AbsoluteValue{\triangledown_x\mathbf{I}(x,y)} + \AbsoluteValue{\triangledown_y\mathbf{I}(x,y)}}{2} \leq \gamma_\text{grad}.
\end{align}
Empirically, we found that the gradient loss is more effective for identifying the epoch with the highest performing model than the loss from \Cref{eq:loss}.

\begin{figure}[t]
	\centering
	\begin{subfigure}[]{0.49\columnwidth}
		\includegraphics[width=\columnwidth]{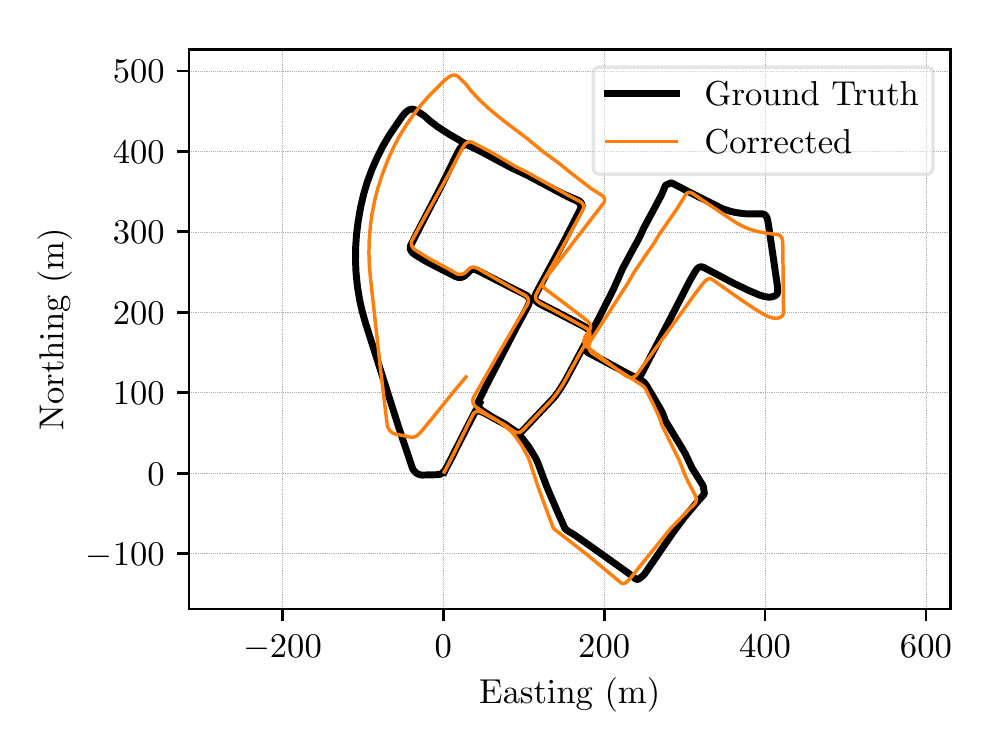}
		\caption{Epoch 5: 40 loop closures detected.}
		\label{fig:mono_online_00}
	\end{subfigure}
	\hfill
	\begin{subfigure}[]{0.49\columnwidth}
		\includegraphics[width=\columnwidth]{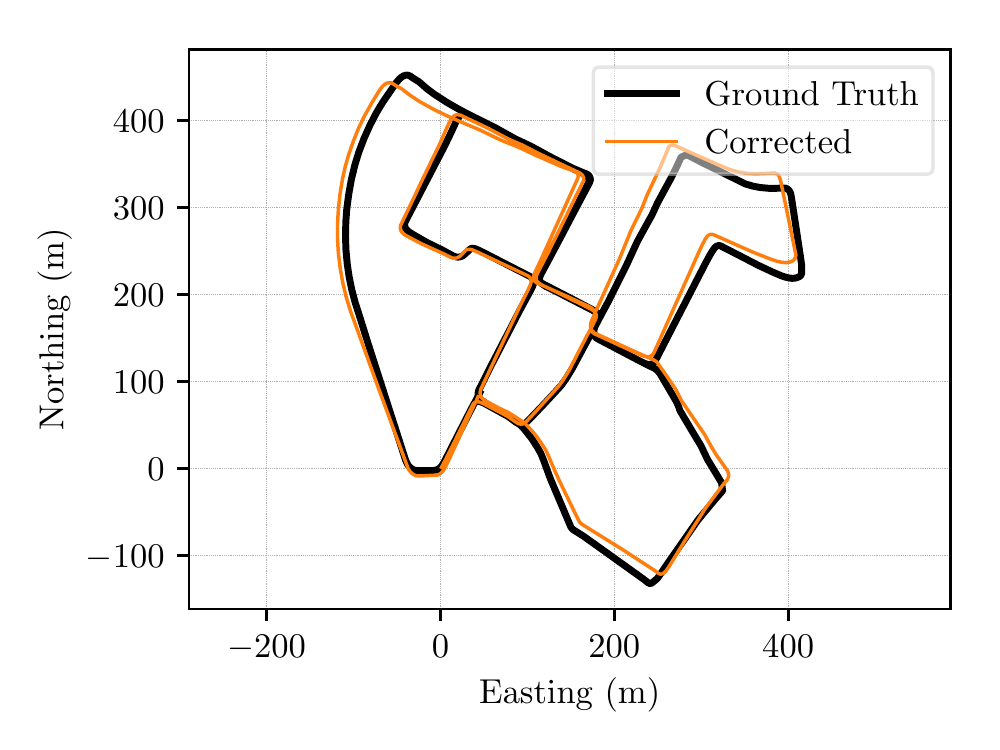}
		\caption{Epoch 28: 122 loop closures detected.}
		\label{fig:mono_online_02}
	\end{subfigure}
	\caption{Loop closure comparison for test sequence (\texttt{00}): compared to ground truth, the epoch with a higher number of detected loop closures resulted in a more accurate trajectory estimate.}
	\label{fig:loopclosure}
	\vspace{-0.3cm}
\end{figure}

\textit{Loop Closure Criterion}: Second, we note that we can relate the number of loop closures in the compounded validation trajectory with the accuracy of the pose corrections in the test set (see \Cref{fig:loopclosure}). To identify loop closures in the validation trajectory, we use predefined thresholds\footnote{For a given pose, $\Transform_{t,0}$, we defined a loop closure event at time $t+n$ if the pose $\mathbf{T}_{t+n,0}$ was within 7 m and $8.5^{\circ}$ of the pose at time $t$. We also ensured that the forward translation of $\Transform_{t+n,t}$ exceeded 10 m to ensure that the vehicle was performing a second traverse along the same path.} and save the model at the training epoch that produces the most  loop closures for the validation set. We emphasize that although this method does require the \textit{validation sequence} to have loop closures (which we ensured herein by selecting the KITTI validation sequences \texttt{00} and \texttt{05} that consist of multiple traverses of the same road) it places no such restriction on the \textit{test sequence}. Therefore, as long as one can find a validation sequence that has multiple traverses of the same path, this epoch selection criterion can be applied.

\begin{table}
	\footnotesize
	\centering
	\caption{Results of correcting \texttt{libviso2-m} with our DPC network. We benchmark against several state-of-the-art learning-based monocular VO estimators.}
	\label{tab:mono_online_traj}
	\begin{threeparttable}
		\begin{tabular}{c@{\hspace{0.2\tabcolsep}}c@{\hspace{0.2\tabcolsep}}c@{\hspace{0.6\tabcolsep}}c@{\hspace{0.9\tabcolsep}}c@{\hspace{0.9\tabcolsep}}c}
			\toprule
			\textbf{Estimator} & \textbf{Stopping Criterion} & \multicolumn{4}{c}{\textbf{Mean Segment Error}} \\ \midrule
			&  & \multicolumn{2}{c}{\textbf{Seq. 09}} & \multicolumn{2}{c}{\textbf{Seq. 10}} \\ \cmidrule{3-6} 
			\textbf{} &  & \begin{tabular}[c]{@{}c@{}}Trans. \\ (\%)\end{tabular} & \begin{tabular}[c]{@{}c@{}}Rot. \\ ($^o$/100m)\end{tabular} & \begin{tabular}[c]{@{}c@{}}Trans. \\ (\%)\end{tabular} & \begin{tabular}[c]{@{}c@{}}Rot. \\ ($^o$/100m)\end{tabular} \\ \midrule
			\texttt{libviso2-m} & & 8.66 & 2.76 & 8.00 & 3.32 \\
			SfMLearner \cite{zhou:2017}  & --- & 18.8 & 3.21 & 14.3 & 3.30 \\
			UnDeepVO \cite{Li:2018} & --- & 7.01 & 3.61 & 10.6 & 4.65 \\
			Zhan et al. \cite{Zhan:2018} & --- & 11.9 & 3.60 & 12.6 & 4.65 \\
			Zhu et al. \cite{Zhu:2018} & --- & 4.66 & 1.69 & 6.30 & 1.59 \\
			Luo et al. \cite{Luo:2018} & --- & 3.72 & 1.60 & 6.06 & 2.22 \\
			Ours$^1$ & \textit{Gradient Loss} & 2.82 & \textbf{0.76} & 3.81 & \textbf{1.34} \\
			& \textit{Loop Closure} & \textbf{2.13} & 0.80 & \textbf{3.48} & 1.38 \\ \bottomrule	
		\end{tabular}
		\begin{tablenotes}
			\item[1] Validation sequence \texttt{05} for sequence \texttt{09}, and \texttt{00} for sequence \texttt{10}.
		\end{tablenotes}
	\end{threeparttable}
\vspace{-0.3cm}
\end{table}

\begin{figure}[t]
	\centering
	\begin{subfigure}[]{0.48\textwidth}
		\centering
		\includegraphics[width=0.85\textwidth]{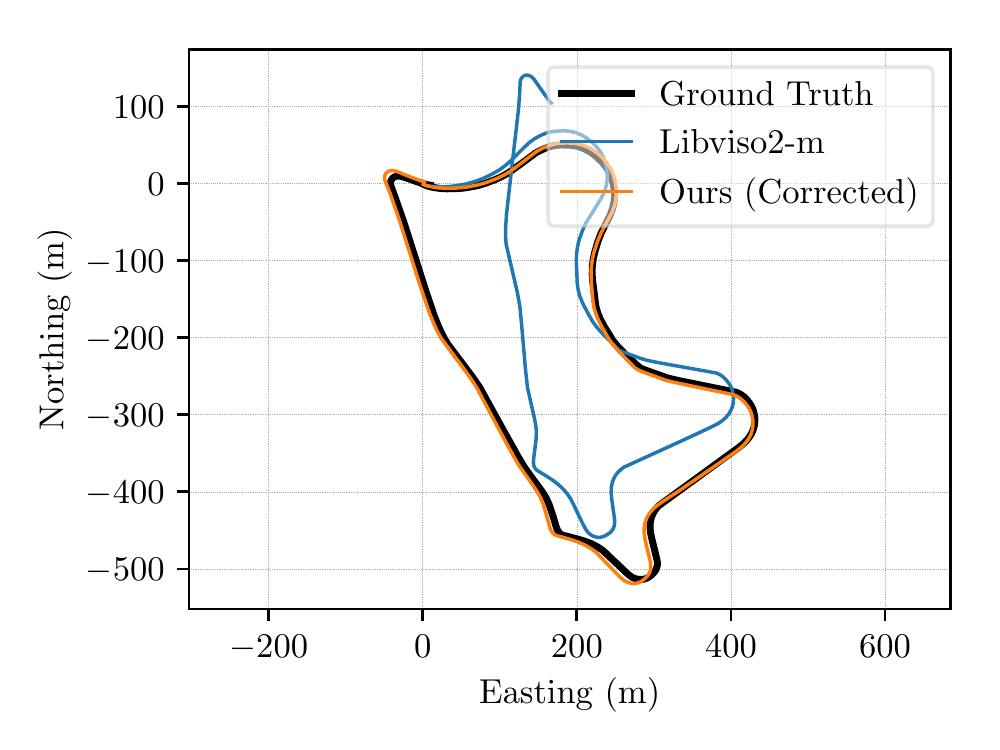}
		\vspace{-0.6cm}
		\label{fig:lc1}
	\end{subfigure}
	\begin{subfigure}[]{0.48\textwidth}
		\centering
		\includegraphics[width=0.85\textwidth]{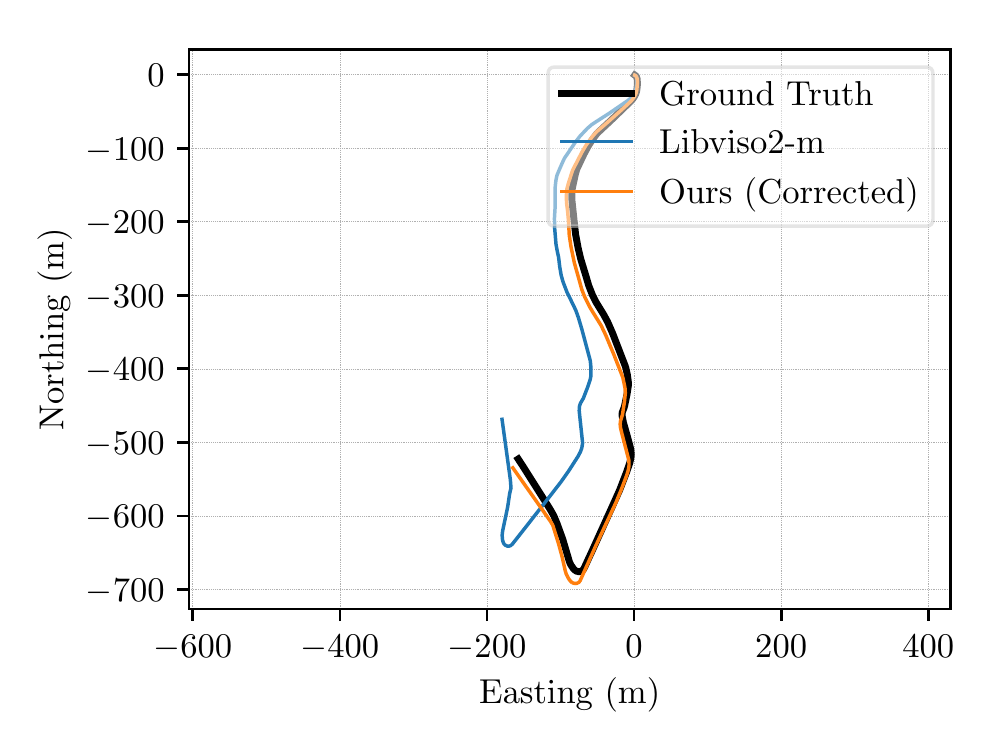}
		\label{fig:lc2}
	\end{subfigure} 
	\vspace{-0.2cm}
	\caption{Corrected \texttt{libviso2-m} estimates for sequences \texttt{09} (top) and \texttt{10} (bottom). We show the original \texttt{libviso2-m} estimate.}
	\label{fig:mono-traj}
\vspace{-4mm}
\end{figure}

\section{Results}
\label{sec:results}

Our evaluation metrics were the mean absolute trajectory error (m-ATE) and the mean segment error (m-SE). In accordance with the KITTI odometry benchmark, we computed the mean error for all segments that were $\{100,200,\cdots, 800\}$ metres in length; we report the overall average herein.

\begin{table*}[]
	\footnotesize
	\centering
	\caption{Results of correcting \texttt{libviso2-s} with our self-supervised DPC network.}
	\label{tab:stereo_results}
	\begin{threeparttable}
		\begin{tabular}{ccccccc}
			\toprule
			&  &  & \multicolumn{2}{c}{\textbf{m-ATE}} & \multicolumn{2}{c}{\textbf{m-SE}} \\ \cmidrule{4-7} 
			\textbf{\begin{tabular}[c]{@{}c@{}}Test Sequence \\ (Length)\end{tabular}} & \textbf{Estimator} & \begin{tabular}[c]{@{}c@{}}\textbf{Stopping} \\ \textbf{Criterion} \end{tabular} & \begin{tabular}[c]{@{}c@{}}Trans. \\ (m)\end{tabular} & \begin{tabular}[c]{@{}c@{}}Rot.\\ ($^\circ$)\end{tabular} & \begin{tabular}[c]{@{}c@{}}Trans.\\  (\%)\end{tabular} & \begin{tabular}[c]{@{}c@{}}Rot. \\ ($^\circ$/100m)\end{tabular} \\ \midrule
			\texttt{00} (3.7 km) & \texttt{libviso2-s} & --- & 53.77 & 13.30 & 2.79 & 1.292 \\
			& \texttt{libviso2-s} + \DPCNet{} \cite{Peretroukhin:2018} & --- & 15.68 & 3.07 & 1.62 & 0.559 \\
			& Direct Keyframe & --- & \textbf{12.41} & \textbf{2.45} & 1.28 & 0.542 \\
			& Ours$^1$ & \textit{Gradient Loss} & 12.59 & 2.47 & \textbf{0.99} & 0.457 \\
			&  & \textit{Loop Closure} & 14.65 & 3.32 & 1.03 & \textbf{0.444} \\ \midrule
			\texttt{02} (5.1 km) & \texttt{libviso2-s} & --- & 68.60 & 12.55 & 2.42 & 0.923 \\
			& \texttt{libviso2-s} + \DPCNet{} & --- & 17.69 & 2.86 & 1.16 & 0.436 \\
			& Direct Keyframe & --- & 16.33 & 3.19 & 1.21 & 0.467 \\
			& Ours$^1$ & \textit{Gradient Loss} & \textbf{15.69} & 3.52 & 1.11 & 0.499 \\
			&  & \textit{Loop Closure} & 21.31 & \textbf{1.91} & \textbf{0.83} & \textbf{0.373}\\ \midrule
			\texttt{05} (2.2 km) & \texttt{libviso2-s} & --- & 19.68 & 6.30 & 2.31 & 1.135 \\
			& \texttt{libviso2-s} + \DPCNet{}  & --- & 9.82 & 3.57 & 1.34 & 0.562 \\
			& Direct Keyframe & --- & 5.83 & 2.05 & \textbf{0.69} & 0.320 \\
			& Ours$^2$ & \textit{Gradient Loss} & 10.92 & 4.10 & 1.33 & 0.597 \\
			&  & \textit{Loop Closure} & \textbf{4.03} & \textbf{1.18} & 0.83 & \textbf{0.304}\\ \bottomrule
		\end{tabular}
		\begin{tablenotes}
			\item[1] Validation sequence \texttt{05}.$^2$ Validation sequence \texttt{00}.  
		\end{tablenotes}
	\end{threeparttable}
\vspace{2mm}
\end{table*}

\begin{figure*}[]
	\centering
	\begin{subfigure}[]{0.32\textwidth}
		\includegraphics[width=\columnwidth]{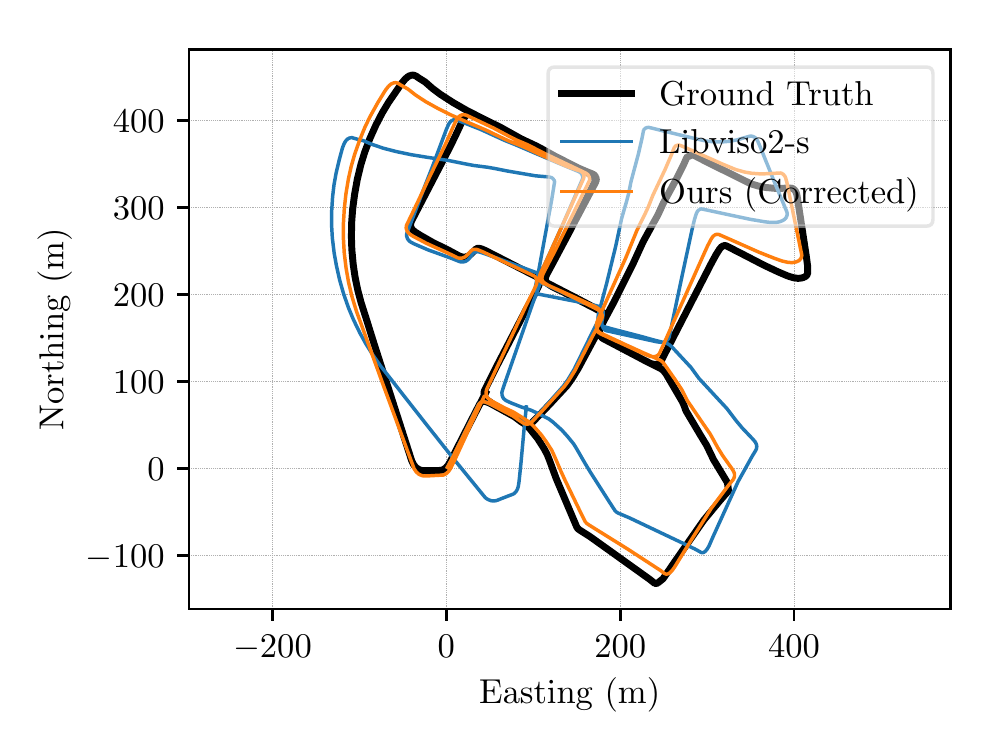}
		\caption{Sequence \texttt{00}}
		\label{fig:opt00}
	\end{subfigure}
	\hfill
	\begin{subfigure}[]{0.32\textwidth}
		\includegraphics[width=\textwidth]{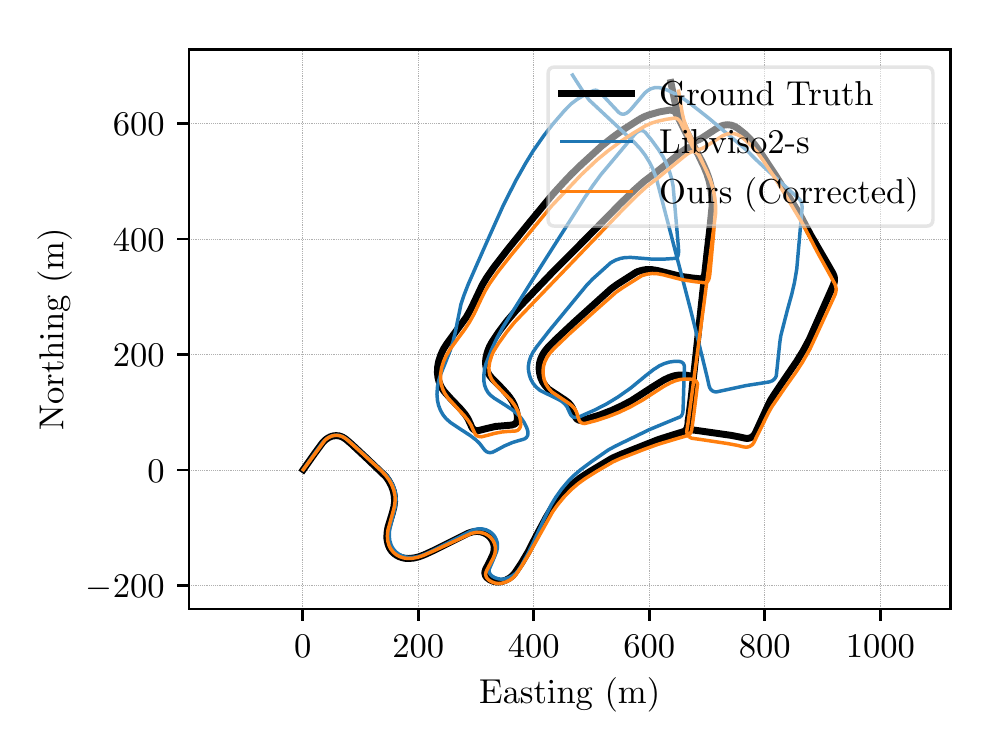}
		\caption{Sequence \texttt{02}}
		\label{fig:opt02}
	\end{subfigure}
	\hfill
	\begin{subfigure}[]{0.32\textwidth}
		\includegraphics[width=\textwidth]{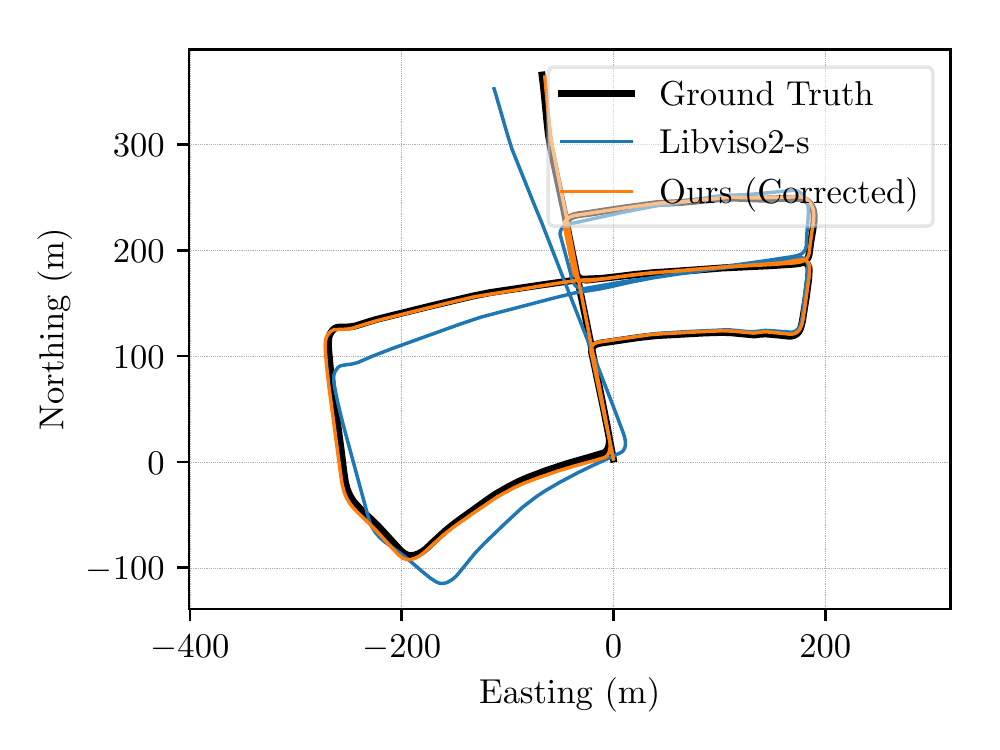}
		\caption{Sequence \texttt{05}}
		\label{fig:opt05}
	\end{subfigure}
	\caption{Corrected \texttt{libviso2-s} trajectories. We show the original \texttt{libviso2-s} estimate for comparison.}
	\label{fig:stereo_online_traj}
\vspace{-2mm}
\end{figure*}

\textit{Monocular Results}: We applied corrections to the \texttt{libviso2-m} estimates and compared the corrected trajectories with state-of-the-art end-to-end monocular VO estimators. Prior to applying corrections, we rescaled the \texttt{libviso2-m} estimate to be consistent with ground truth by adjusting each relative pose change to match the ground truth inter-frame displacement (while this is not fully self-supervised, this is in line with how other monocular VO systems \cite{zhou:2017,Zhan:2018} are evaluated). \Cref{tab:mono_online_traj} shows the mean segment errors for test sequences \texttt{09} and \texttt{10} of the KITTI dataset, which are the most common test sequences for monocular methods; notably, we achieve state-of-the-art accuracies compared to modern learning-based techniques. \Cref{fig:mono-traj} illustrates the corrected trajectories, which appear to be significantly more accurate than the original \texttt{libviso2-m} estimate. 

\textit{Stereo Results}: We applied corrections to \texttt{libviso2-s} and compared the corrected trajectories with the corrected trajectories from \DPCNet{} \cite{Peretroukhin:2018}. In practice, we found that only rotation ($\LieGroupSO{3}$) corrections were required to adjust the \texttt{libviso2-s} estimates, since the translation estimates were already highly accurate. Our DPC network uses monocular images only to regress pose corrections and so cannot provide any improvement by correcting the translation estimates. We evaluated our DPC network on test sequences \texttt{00}, \texttt{02}, \texttt{05}. \Cref{tab:stereo_results} lists the m-ATE and m-SE for our corrected trajectories, while \Cref{fig:stereo_online_traj} visually depicts these results and compares them to the \texttt{libviso2-s} estimates. We benchmark against \DPCNet{} \cite{Peretroukhin:2018} and also a direct, keyframe-based VO implementation based on DSO \cite{Engel-et-al-pami2018}. Both of our proposed modes of operation are consistently more accurate than these competing methods for the three test sequences.

\section{Conclusions and Future Work}

In this paper, we presented a deep network that is trained to correct classical VO estimators in a self-supervised manner, without the need for 6-DoF ground truth. By regressing pose corrections (instead of the full inter-frame pose change), our approach produces trajectory estimates that are significantly more accurate than existing state-of-the-art end-to-end VO networks. We attribute this increase in accuracy to the union of learning-based and classical (handcrafted) models. Our method preserves the core geometric framework that generally yield accurate egomotion estimates under nominal conditions, and pairs it with a learning approach that applies corrections when modelling assumptions are violated or other confounding factors are present. Our self-supervised loss formulation facilitates continual model retraining with new data. As future work, we plan to incorporate stereo constraints (e.g., the left-right consistency constraint of \cite{Li:2018}) or other sources of metric information (e.g., inertial measurement unit data) to better improve our translation corrections. 

\section*{Acknowledgments}

This work was supported in part by the Natural Sciences and Engineering Research Council (NSERC) of Canada. We gratefully acknowledge the contribution of NVIDIA Corporation, who provided the Titan X GPU used for this research through their Hardware Grant Program.

\newpage
\balance
\clearpage
\bibliographystyle{IEEEtran}
\bibliography{example.bib}
\end{document}